\documentclass[letterpaper]{article} 
\usepackage{aaai24}  
\usepackage{times}  
\usepackage{helvet}  
\usepackage{courier}  
\usepackage[hyphens]{url}  
\usepackage{graphicx} 
\usepackage{natbib}  
\usepackage{caption} 
\usepackage{algorithm}
\usepackage{algorithmic}
\usepackage{booktabs}
\usepackage{multicol}
\usepackage{multirow}
\usepackage{enumitem}
\usepackage{amsmath}
\usepackage{booktabs}
\usepackage{newfloat}
\usepackage{listings}
\DeclareCaptionStyle{ruled}{labelfont=normalfont,labelsep=colon,strut=off} 
\floatstyle{ruled}
\newfloat{listing}{tb}{lst}{}
\floatname{listing}{Listing}
\newcommand{\n}{{\sf ALISON}}
\usepackage{xcolor}

\title{{\n}: Fast and Effective Stylometric Authorship Obfuscation}

\author{
    Eric Xing\textsuperscript{\rm 1},
    Saranya Venkatraman\textsuperscript{\rm 2},
    Thai Le\textsuperscript{\rm 3},
    Dongwon Lee\textsuperscript{\rm 2}
}
\affiliations{
    \textsuperscript{\rm 1}McKelvey School of Engineering, Washington University in St. Louis, MO, USA\\
    \textsuperscript{\rm 2}College of Information Sciences and Technology, The Pennsylvania State University, PA, USA\\
    \textsuperscript{\rm 3}School of Engineering, University of Mississippi, MS, USA\\

    e.xing@wustl.edu, 
    saranyav@psu.edu,
    thaile@olemiss.edu,
    dongwon@psu.edu
}

\begin{document}

\maketitle

\begin{abstract}
{\it Authorship Attribution} (AA) and {\it Authorship Obfuscation} (AO) are two competing tasks of increasing importance in privacy research. Modern AA leverages an author's consistent writing style to match a text to its author using an AA classifier. AO is the corresponding adversarial task, aiming to modify a text in such a way that its semantics are preserved, yet an AA model cannot correctly infer its authorship. 
To address privacy concerns raised by state-of-the-art (SOTA) AA methods,
new AO methods have been proposed but remain largely impractical to use due to their prohibitively slow training and obfuscation speed, often taking hours.
To this challenge, we propose a practical AO method, {\n}, that 
(1) dramatically reduces training/obfuscation time, demonstrating more than 10x faster obfuscation than SOTA AO methods, 
(2) achieves better obfuscation success through attacking three transformer-based AA methods on two benchmark datasets, typically performing 15\% better than competing methods,
(3) does not require direct signals from a target AA classifier during obfuscation, and 
(4) utilizes unique stylometric features,  allowing sound model interpretation for explainable obfuscation.
We also demonstrate that {\n} can effectively prevent four SOTA AA methods from accurately determining the authorship of ChatGPT-generated texts, all while minimally changing the original text semantics. To ensure the reproducibility of our findings, our code and data are available at:
\url{https://github.com/EricX003/ALISON}.



\end{abstract}

\section{Introduction}
Writing styles are often consistent among texts written by the same author. However, the writing styles of different authors can be very dissimilar. Therefore, the authorship identity of an anonymous piece of writing can still be revealed by analyzing its writing style and matching it to a pool of known authorship markers, a task known as {\bf Authorship Attribution} (AA). In a machine learning context, authorship markers are predictive signals that can distinguish one author's writing style from the others. Such signals are often called stylometric features. 
Multiple types of stylometric features, including lexical features (e.g., structure of words and frequency of different character sequences), syntactic features (e.g., part-of-speech distributions and occurrences of functional words and punctuation), and content features (e.g., semantics of words and phrases in the text) are engineered to allow a machine learning model to match a text to an authorship label.
These engineered features, such as Writeprints~\cite{writeprints}, often include not one but several interpretable signals such as word and character bigrams, word length distributions, or special character frequencies to improve the classification accuracy.

However, recent AA techniques~\cite{fabien-etal-2020-bertaa, devlin-etal-2019-bert} utilize complex transformer models--e.g., BERT~\cite{textfooler}, RoBERTa~\cite{Liu2019RoBERTaAR}, BertAA~\cite{fabien-etal-2020-bertaa}, to automatically learn useful features for AA from raw text. This removes the need to rely on explicitly engineered stylometric features. While these models are more computationally expensive to train and notorious for their lack of interpretability, they significantly outperform traditional AA classifiers ~\cite{fabien-etal-2020-bertaa}.

\begin{figure}[tb!]
    \centering
    \includegraphics[width=0.75\columnwidth]{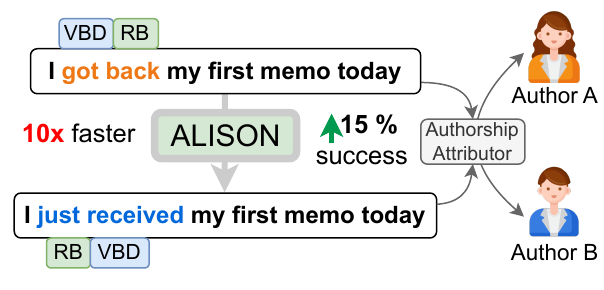}
     \caption{{\n} successfully obfuscating a text by changing its style while preserving semantics.}
    \label{fig:obfex}
\end{figure}

As AA techniques become more accurate and efficient, they are more likely to be exploited by malicious actors to detect authorship identities behind anonymous texts. This is severely detrimental to a number of groups, especially NGO activists, whistleblowers, and journalists. 
As current SOTA transformer-based AA models are sufficiently powerful, it becomes important to develop methods that reduce the risk of an anonymous text's true authorship being exposed. Therefore, in this work, we study the opposite task of AA, known as {\bf Authorship Obfuscation} (AO), which aims to thwart authorship attribution classifiers by making a few changes to the input text in a systematic way. Successful AO will fool the target model into making an incorrect attribution out of a pool of candidates. Because AA techniques generally degrade in performance as the number of authors becomes large (i.e. $>100$), and an adversary can generally narrow the pool of authors down to a small, finite set, we do not consider authorship obfuscation in the open-world setting. Figure~\ref{fig:obfex} shows an example of a successful authorship obfuscation against a BERT-based~\cite{devlin-etal-2019-bert} authorship attributor.

There are three important properties that we desire in a ``practical" AO approach: (1) ability to operate without significant knowledge of the adversary, (2) fast running time for long-form texts ($<1$ second, not minutes or hours), and (3) intuitive interpretability for a trustworthy obfuscation process. 
Unfortunately, SOTA AO methods, do not satisfy these properties at all,
often requiring long running time to obfuscate text in a black-box fashion while making numerous calls to the attacked model. Such methods are impractical because a black-box understanding of the model to be attacked is often impossible to obtain, prohibitive running times diminish the productivity of an author seeking anonymity, and a lack of interpretability during obfuscation prevents current methods from being trustworthy.



To address the aforementioned limitations of current AO methods, we propose a novel stylometry-grounded novel obfuscation method, {\n}: (\emph{F\underline{\textbf{a}}st Sty\underline{\textbf{l}}ometr\underline{\textbf{i}}c Author\underline{\textbf{s}}hip \underline{\textbf{O}}bfuscatio\underline{\textbf{n}}}), which overcomes these challenges as follows:

\begin{itemize}
    \item {\n} significantly reduces the obfuscation runtime by over 10x while also achieving better semantic preservation during obfuscation.
    \item {\n} consistently outperforms competing approaches by around 15\% in obfuscation success rate. 
    \item {\n} is also able to provide explanations for its obfuscation results with interpretable stylometric features.
\end{itemize}

\begin{figure*}[h]
    \centering
    \includegraphics[width=\textwidth]{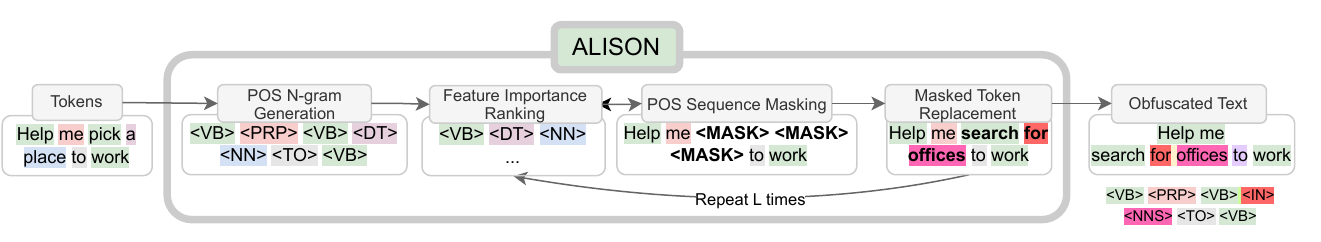}
    \caption{{\n}: Our proposed obfuscation pipeline.}
    \label{fig:MethodDiagram}
\end{figure*}

\section{Background}

We narrow the scope of our work to the blind AO setting where a textual adversarial attack against AA classifier has two main constraints: (1) the attacker cannot query the AA classifier, and (2) the attacker also does not have access to its architecture, training data, etc. These constraints make the AO task more challenging but also more practical than existing threat models often used in existing literature where a public API to the target AA classifier is assumed to be accessible. The following section describes existing work pertinent to this AO setting. 

Mutant-X~\cite{MUTANT} is an automated obfuscation method that utilizes genetic algorithms to iteratively make single-word substitutions by examining the confidence degradation gleaned from a black-box understanding of the attacked model. While this is a black-box attack method, we repurposed it as a blind attack as described in transferability studies associated with its original paper~\cite{MUTANT}. Avengers Ensemble~\cite{Haroon2021AvengersEI} attempts to improve upon Mutant-X and decrease reliance on black-box knowledge of the target classifier by utilizing an ensemble-based internal classifier to improve the transferability of the method to a variety of adversaries, which boosts its performance in the blind attack setting. We will refer to this method as Avengers for the rest of the paper.



 
Other popular greedy-based black-box methods in the NLP adversarial literature, such as TextFooler~\cite{textfooler} and BERT-Attack~\cite{li-etal-2020-BERT-Attack}, often have a high degree of dependence on the accessibility to the target AA classifier they attack. These methods make queries to the victim model per token in order to obtain a logit-based ranking of word importance. Then, top tokens may be replaced with close neighbors in precomputed embedding spaces~\cite{textfooler} or by leveraging token representations of large language models~\cite{li-etal-2020-BERT-Attack}. However, these methods often demonstrate a sharp decline in performance once the attacks are transferred to different target classifiers~\cite{textfooler}. Additionally, such methods generally lack interpretability, as model explanations are based solely on the black-box model that is being attacked instead of revealing identifying linguistic patterns. 

Lastly, large generative language models, such as ChatGPT~\cite{instructgpt}, have demonstrated impressive paraphrasing capability which may be suitable for AO applications. A user may obtain a stylometrically different but semantically consistent text by prepending a fixed paraphrasing prompt to query a language model.
 

\section{Problem Formulation}
Given a text corpus $\mathcal{X}$, we define an AA classifier $f$ trained on $\mathcal{X}$, such that for arbitrary text $x\in \mathcal{X}$, $f(x)$ attributes the authorship of $x$. Given $\mathcal{T}$ is a set of texts to obfuscate, our objective is to thwart $f$ for any text $t\in \mathcal{T}$ by transforming $t$ into $t'$ such that $f(t){\neq}f(t')$. We assume that $\mathcal{X}$ and $\mathcal{T}$ share the same pool of potential authors and are in a similar domain--e.g., news articles, blog posts-- but do not contain any identical texts.

Moreover, we also assume no access to $\mathcal{X}$ by the adversary. However, they do have access to another non-overlapping corpus $\mathcal{X}^*$ with a similar size containing the same pool of authors and domain with $\mathcal{X}$. Such assumption is reasonable in practice, especially when online social networks have made it very convenient for anyone to access text content generated by millions of people worldwide. To evaluate our approach in this setting, we split each publicly available text classification corpus into three disjoint sets, $\mathcal{X}$, $\mathcal{X}^*$, and $\mathcal{T}$ stratified by unique authorship labels. 


\section{Proposed Method}
Figure \ref{fig:MethodDiagram} illustrates {\n}'s overall obfuscation pipeline. {\n} is designed to reduce computational complexity while advancing obfuscation success and semantic preservation during obfuscation. To do this, we employ three overarching strategies. First, we train an internal, lightweight AA classifier \textit{once} that uses intuitive linguistic properties of part-of-speech (POS) sequences to guide the obfuscation process. Second, we aim to obfuscate a phrase of multiple words at a time instead of perturbing token by token. Third, we leverage an advanced pre-trained language model (PLM) to generate the replacement token sequence that best fits the sentence context and semantics without making queries to an embedding space.

\subsection{\textit{One-Time} Stylistic Internal AA Classifier Training} \label{sec:stylistic_aa}

Because blind attacks on AA models often rely on an internal approximation of an arbitrary adversarial classifier to choose candidate words or phrases to be replaced, tuning the internal classifier for maximal transferability to other target classifiers is integral to producing high obfuscation success rate~\cite{Haroon2021AvengersEI}. Therefore, we augment the traditional internal classifier feature space of character n-grams with POS n-grams, features we believe to be more heavily rooted in true style. We hypothesize that while writing style encompasses word and character frequencies, more generally, writing style also encompasses frequencies of individual POS tags and their collocations. Intuitively, POS and sequences of several POS tags capture writing style because they do not describe the content of the text but rather how the ideas in the text are synthesized. Generally, an author's texts should contain similar POS sequence patterns, as they represent common textual structures used to synthesize different ideas. 

\subsubsection{Feature Extraction.} 
\begin{figure}[!h]
    \includegraphics[width=\columnwidth]{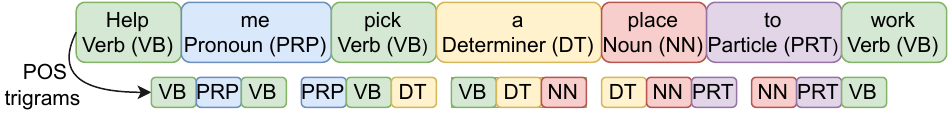}
    \caption{An example of extracting POS trigrams.}
    \label{fig:posngram_diagram}
\end{figure}

We first extract the POS tags of all texts in the corpus $\mathcal{X}^*$. Next, we extract character and POS tag n-grams of various lengths as features for training the internal classifier. 
Figure \ref{fig:posngram_diagram} demonstrates the procedure of extracting POS n-grams from a sample sentence with $n{\leftarrow}3$.

An n-gram is a contiguous sequence of $n$ linguistic units (e.g., characters, words, POS tags) within a text. Given a set of sequence lengths $V$, for each length $l \in V$, we extract all character and POS level $l$-grams over the entire training corpus and collect the $L$ most frequent character and POS $l$-grams. The normalized frequencies of these $L$ most frequent character and POS $l$-grams for each length $l \in V$ are concatenated to form the stylistic representations of the text.

\subsubsection{Internal Classifier Training.} The resulting vector representations are then used to train a fully connected neural network (NN) model on the authorship attribution task. We opt for a simple NN due to its computational efficiency without much compromise on generalization. To utilize this model for prioritizing which phases or words in a sentence to perturb first, we then extract a list of features, ranked by importance, for $\forall t \in T$ using Integrated Gradients ~\cite{sundararajan2017axiomatic}, a model interpretability algorithm that assigns an importance score to each input feature by approximating the integral of the gradients with respect to the input.

 
We also multiply each extracted importance by the term $c^{length(feature)}$ for each feature's attribution, where $c$ is a constant. During experimentation, we empirically observed that shorter POS n-grams were more abundant at the beginning of the attribution-ranked n-gram lists. We believe that this behavior is because of the necessarily lesser frequency of an arbitrarily longer POS n-gram in typical texts, as each longer n-gram occurrence necessarily is an occurrence of all contiguous substrings of the n-gram, i.e., shorter n-grams. Therefore, we introduced this scaling constant to artificially inflate the importance of longer POS n-grams to compensate for this behavior.

\subsection{Replacement Phrase Generation via Masked PLM} 
To perform obfuscation, we must be able to generate replacement phrases using existing phrases as prompts. To do this, we leverage the masked language modeling approach used by ~\citet{devlin-etal-2019-bert}. More specifically, given a sentence and the desired word tokens to be replaced, we mask the tokens to be replaced and use this modified text as input for a BERT model under a masked token prediction task. The top prediction for each masked token is used as the word's replacement. By using a SOTA language model, we aim to minimize the degree of information loss, as the language model will be able to infer much of the contents of the phrase through context but may scramble POS sequences, which hides authorship. This token-sequence masking procedure lies at the core of {\n}'s speed-up, allowing a single PLM forward pass to perturb multiple tokens.

\subsection{Text Obfuscation Process: One N-Gram at a Time} 
To obfuscate each $t \in T$, we first extract the POS tags and n-gram features for $t$, which are used to compute importance values as described previously. Then, we iterate through the ranked feature list in descending order of importance, omitting character n-gram features (only considering POS n-gram features) and pick the top $L$ features.  We omit character n-grams because important character n-grams are generally functional words or involve punctuation, which would negatively impact fluency upon perturbation.

Next, we attempt to match each of the top $L$ POS n-grams to the POS n-gram profile of $t$. For each n-gram match found, we update $t$ through the phrase generation procedure as described previously. Lastly, we mark this phrase as changed so that it cannot be changed in subsequent steps as to prevent any specific section of text from deviating significantly from the original. Obfuscation is complete once all matches for the top $L$ POS n-grams are processed. 

One unique property of {\n} is that it will modify the text even if the internal classifier believes it will be classified incorrectly. This property is desirable because {\n} will uniformly obfuscate all texts, likely decreasing adversarial classifier confidence even if a complete obfuscation is unsuccessful. This differs from logit query-based methods because they do not attempt to perform \emph{any} obfuscation if their internal classifier's prediction does not match the ground truth, leading to a large proportion of $t\in \mathcal{T}$ being completely unedited and therefore vulnerable.

\section{Experimental Setup}

\noindent \textbf{Datasets.} We use \textit{TuringBench}~\cite{Uchendu2021TuringBenchAB} to evaluate {\n} on machine-generated texts. TuringBench is a collection of 160K human and machine-generated texts across 20 authors, 19 of which are neural text generation models, and one of whom is human. 
We also use the \textit{Blog Authorship Corpus}~\cite{BLOG} to evaluate {\n} on human-written texts. The dataset consists of the aggregated blog posts of 19,320 bloggers gathered from blogger.com, of which we select only the blogs from the top-10 most frequent authors. Both datasets are publicly available. We report all AO results on the test set.

\vspace{3pt}
\noindent \textbf{Target Classifiers.} We use three SOTA transformer-based models as target AA classifiers to attack: BERT~\cite{devlin-etal-2019-bert}, DistilBERT~\cite{sanh2019distilbert}, and RoBERTa~\cite{Liu2019RoBERTaAR}. 
These adversarial classifiers were trained on the \textit{1st disjoint half} of the training and validation sets. They achieved around 80\% testing accuracy on on TuringBench, while demonstrating varying performance on the Blog Authorship Corpus, ranging from approximately 85\% (DistilBERT) to 95\% (RoBERTa) testing accuracy. 

\vspace{3pt}
\noindent \textbf{Obfuscation Baselines and Internal Classifier Training.} We utilize TextFooler, Mutant-X, Avengers, BERT-Attack, and ChatGPT as baselines to compare against our proposed AO framework {\n}. Except for ChatGPT, these methods all maintain an internal classifier for reference during obfuscation. While many of these are black-box attack methods, we repurposed them for the blind attack setting using the internal classifier specifications given in transferability studies instead of giving them access to our SOTA target models. Our neural-network-based n-gram classifier is trained on the disjoint 2nd half of the training and validation data that was not used to train our SOTA target models using $V=\{1,2,3,4\}$. Internal classifiers for Mutant-X and Avengers were trained as outlined by their papers \cite{Haroon2021AvengersEI, MUTANT} on the same data as our internal classifier. 
TextFooler was trained with both word-based CNN (wordCNN)~\cite{kim2014convolutional} and word-based LSTM (wordLSTM) internal classifiers as specified in their public implementation. We additionally tested TextFooler using our n-gram-based NN model (denoted as TextFooler-POS) to provide a fair comparison and illustrate the effectiveness of our stylometry-grounded approach. BERT-Attack was trained using standard BERT~\cite{devlin-etal-2019-bert}. ChatGPT-based obfuscation was performed by pretending a fixed paraphrasing prompt to each text and obtaining the returned machine response.

\vspace{3pt}

\renewcommand{\tabcolsep}{1.5pt}
\begin{table*}[!t]
    \centering
    \footnotesize
    \begin{tabular}{lccccc}
        \toprule
        \multicolumn{1}{l}{\multirow{4}{*}{\textbf{Method}}} &
        \multicolumn{2}{c}{\textbf{Obfuscation Success (Lower is Better)}} &
        \multicolumn{3}{c}{\textbf{Semantic Preservation (Higher is Better)}} \\ \cmidrule{2-6} 
        \multicolumn{1}{l}{} & \textbf{Accuracy}$\downarrow$ & \textbf{F1-Score}$\downarrow$ & \textbf{METEOR}$\uparrow$ & \textbf{USE Cosine Similarity}$\uparrow$ & \textbf{BERTScore}$\uparrow$ \\ 
        
        \cmidrule(lr){2-3}\cmidrule(lr){4-6}
        \multicolumn{6}{c}{\textbf{TuringBench}} \\ 
        \hline
        \multicolumn{6}{c}{BERT}  \\ \hline
        
        \multicolumn{1}{l}{Mutant-X} & 0.8987 & 0.8798 & 0.8381 & 0.9159 & 0.9366 \\
        \multicolumn{1}{l}{Avengers} & 0.8354 & 0.8334 & 0.8333  & 0.9030 & 0.9320 \\
        \multicolumn{1}{l}{TextFooler-wordCNN} & 0.7089 & 0.6797 & 0.8667 & 0.9614 & 0.9386 \\
        \multicolumn{1}{l}{TextFooler-wordLSTM} & 0.7342 & 0.6935 & \textbf{0.8813} & 0.9671 & 0.9430 \\
        \multicolumn{1}{l}{TextFooler-POS} & 0.7595 & 0.7011 & 0.8650 & 0.9635 & 0.9382 \\
        \multicolumn{1}{l}{BERT-Attack}  & 0.9114 & 0.9179 & 0.8388 & 0.8701 & 0.9526 \\
        \multicolumn{1}{l}{ChatGPT} & 0.7089 & 0.6566 & 0.8373 & 0.9113 & 0.9490 \\
        \multicolumn{1}{l}{{\n}} & \textbf{0.6962 (-1.79\%)} & \textbf{0.6065 (-7.63\%)} & 0.8505 (-3.49\%) & \textbf{0.9682 (0.11\%)} & \textbf{0.9583 (0.60\%)} \\
        
        \hline
        \multicolumn{6}{c}{DistilBERT}  \\ 
        \hline
        
        \multicolumn{1}{l}{Mutant-X} & 0.9494 & 0.9464 & 0.8450 & 0.9192 & 0.9406 \\
        \multicolumn{1}{l}{Avengers} & 0.9113 & 0.8515 & 0.8341  & 0.9048 & 0.9320 \\
        \multicolumn{1}{l}{TextFooler-wordCNN} & 0.7848 & 0.7556 & 0.8641 & 0.9609 & 0.9413 \\
        \multicolumn{1}{l}{TextFooler-wordLSTM} & 0.7722 & 0.7705 & \textbf{0.8819} & 0.9677 & 0.9447 \\
        \multicolumn{1}{l}{TextFooler-POS} & 0.7972 & 0.7955 & 0.8675 & 0.9657 & 0.9391 \\
        \multicolumn{1}{l}{BERT-Attack}  & 0.8228 & 0.8172 & 0.8434 & 0.8737 & 0.9538 \\
        \multicolumn{1}{l}{ChatGPT} & 0.7456 & 0.6474 & 0.8428 & 0.9142 & 0.9494 \\
        \multicolumn{1}{l}{{\n}} & \textbf{0.5823 (-21.90\%)} & \textbf{0.4925 (-23.93\%)} & 0.8538 (-3.19\%) & \textbf{0.9685 (0.08\%)} & \textbf{0.9588 (0.52\%)} \\
        
        \hline
        \multicolumn{6}{c}{RoBERTa} \\ 
        \hline
        
        \multicolumn{1}{l}{Mutant-X} & 0.9014 & 0.8527 & 0.8182 & 0.9062 & 0.9306 \\
        \multicolumn{1}{l}{Avengers} & 0.8028 & 0.7393 & 0.8157  & 0.8967 & 0.9248 \\
        \multicolumn{1}{l}{TextFooler-wordCNN} & 0.6901 & 0.6074 & 0.8621 & 0.9618 & 0.9386 \\
        \multicolumn{1}{l}{TextFooler-wordLSTM} & 0.7606 & 0.6682 & \textbf{0.8814} & 0.9686 & 0.9446 \\
        \multicolumn{1}{l}{TextFooler-POS} & 0.7606 & 0.6760 & 0.8623 & 0.9624 & 0.9402 \\
        \multicolumn{1}{l}{BERT-Attack}  & 0.8451 & 0.8412 & 0.8279 & 0.8603 & 0.9484 \\
        \multicolumn{1}{l}{ChatGPT} & 0.7924 & 0.6569 & 0.8268 & 0.9057 & 0.9436 \\
        \multicolumn{1}{l}{{\n}}  & \textbf{0.6620 (-4.07\%)} & \textbf{0.5624 (-7.41\%)} & 0.8554 (-2.95\%) & \textbf{0.9701 (0.15\%)} & \textbf{0.9595 (1.17\%)} \\
        
        \hline
        \multicolumn{6}{c}{\textbf{Blog Authorship Corpus}}  \\ 
        \hline
        \multicolumn{6}{c}{BERT}  \\ 
        \hline
        
        \multicolumn{1}{l}{Mutant-X} & 0.9130 & 0.9180 & 0.8325  & 0.8514 & 0.9237 \\
        \multicolumn{1}{l}{Avengers} & 0.9565 & 0.9528 & 0.8894 & 0.9028 & 0.9316 \\
        \multicolumn{1}{l}{TextFooler-wordCNN} & 0.9348 & 0.9305 & 0.8854 & 0.9472 & 0.9356 \\
        \multicolumn{1}{l}{TextFooler-wordLSTM} & 0.9565 & 0.9531 & 0.8811 & 0.9439 & 0.9382 \\
        \multicolumn{1}{l}{TextFooler-POS} & 0.9348 & 0.9476 & 0.8838 & 0.9453 & 0.9321 \\
        \multicolumn{1}{l}{BERT-Attack} & 0.9130 & 0.8914 & \textbf{0.9007} & 0.9221 & 0.9202 \\
        \multicolumn{1}{l}{ChatGPT} & 0.9022 & 0.8908 & 0.6720 & 0.8827 & 0.9368 \\
        \multicolumn{1}{l}{{\n}} & \textbf{0.8804 (-2.42\%)} & \textbf{0.7860 (-11.76\%)} & 0.8296 (-7.89\%) & \textbf{0.9551 (0.83\%)} & \textbf{0.9386 (0.04\%)} \\
        
        \hline
        \multicolumn{6}{c}{DistilBERT}  \\ 
        \hline
        
        \multicolumn{1}{l}{Mutant-X} & 0.9048 & 0.9128 & 0.8209  & 0.8497 & 0.9135 \\
        \multicolumn{1}{l}{Avengers} & 0.9405 & 0.9435 & 0.8826 & 0.9044 & 0.9305 \\
        \multicolumn{1}{l}{TextFooler-wordCNN} & 0.8810 & 0.8570 & 0.8839 & 0.9465 & 0.9356 \\
        \multicolumn{1}{l}{TextFooler-wordLSTM} & 0.8810 & 0.8425 & 0.8786 & 0.9427 & 0.9382 \\
        \multicolumn{1}{l}{TextFooler-POS} & 0.8810 & 0.8591 & 0.8832 & 0.9442 & 0.9349 \\
        \multicolumn{1}{l}{BERT-Attack} & 0.9048 & 0.8784 & \textbf{0.9026} & 0.9245 & 0.9205 \\
        \multicolumn{1}{l}{ChatGPT} & 0.9762 & 0.9712 & 0.6524 & 0.8820 & 0.9347 \\
        \multicolumn{1}{l}{{\n}} & \textbf{0.7738 (-12.17\%)} & \textbf{0.7189 (-14.67\%)} & 0.8431 (-6.59\%) & \textbf{0.9595 (1.37\%)} & \textbf{0.9387 (0.05\%)} \\
        
        \hline
        \multicolumn{6}{c}{RoBERTa} \\ 
        \hline
        
        \multicolumn{1}{l}{Mutant-X} & 0.9895 & 0.9886 & 0.8285  & 0.8514 & 0.9232 \\
        \multicolumn{1}{l}{Avengers} & 1.00 & 1.00 & 0.8886 & 0.9033 & 0.9331  \\
        \multicolumn{1}{l}{TextFooler-wordCNN} & 0.3579 & 0.3397 & 0.8872 & 0.9496 & 0.9370 \\
        \multicolumn{1}{l}{TextFooler-wordLSTM} & 0.3684 & 0.3394 & 0.8832 & 0.9464 & 0.9382 \\
        \multicolumn{1}{l}{TextFooler-POS} & 0.3369 & 0.3295 & 0.8654 & 0.9417 & 0.9339 \\
        \multicolumn{1}{l}{BERT-Attack} & 0.9053 & 0.8737 & \textbf{0.9018} & 0.9239 & 0.9205 \\
        \multicolumn{1}{l}{ChatGPT}  & 0.5684 & 0.5939 & 0.6682 & 0.8844 & 0.9368 \\
        \multicolumn{1}{l}{{\n}} & \textbf{0.3053 (-9.38\%)} & \textbf{0.2912 (-11.62\%)} & 0.8288 (-8.09\%) & \textbf{0.9544 (0.51\%)} & \textbf{0.9452 (0.75\%)} \\
        
        \bottomrule
    \end{tabular}
    \caption{Results from main obfuscation trials, $15<L<25$. Best performance is shown in boldface. The percentage (\%) indicates the performance gain of {\n}  compared to the 2nd best competition if positive (or drop if negative) per each metric.}
    \label{table:attk_success_and_sim}
\end{table*}

\vspace{3pt}
\noindent \textbf{Evaluation Metrics}
\begin{itemize}[leftmargin=\dimexpr\parindent-0.7\labelwidth\relax,noitemsep]
    \item \textbf{Obfuscation Success.} The most intuitive measure of obfuscation success is measuring the target AA model's accuracy. Because there is a potential for the label distribution to become skewed during the removal of misclassified samples, we also measure F1-Score, a more robust metric in such a setting. To analyze the obfuscation success, we also monitor the reduction in target model accuracy between the original and obfuscated texts. Because we only retain correctly classified samples for obfuscation, the baseline accuracy and F1-Score are 1.00. A smaller post-obfuscation accuracy and F1-Score indicates a more successful attack, and therefore greater obfuscation success.
    \item \textbf{Running Time.} First, we recorded the running time of each algorithm, as an obfuscation method that requires a prohibitive amount of resources or computation time may not be scalable to real world AO scenarios. We split this time measurement into two phases, the time associated with one-time training of internal classifiers, and the time associated with the average inference time of the retained samples.
    \item \textbf{Semantic Preservation.} We also measure metrics of semantic preservation or semantic similarity between the original and obfuscated texts. Metrics indicating higher semantic preservation are favorable, as they indicate that there was a limited degree of information loss and that the perturbations to the text would not significantly impair a reader's understanding of the original text. These metrics include (1) \textit{METEOR Score:} METEOR score is a standard for measuring the similarity between two texts in a natural language setting. It is grounded in the measure of alignments of word unigrams among texts; (2) \textit{USE Cosine Similarity:} The Universal Sentence Encoder (USE)~\cite{cer-etal-2018-universal} is a text embedding model that is frequently adopted to accurately capture the semantics of a sentence. We utilize \textit{cosine-similarity} to determine the degree of similarity between generated embeddings; (3) \textit{BERTScore:} BERTScore~\cite{bertscore} is another metric of semantic similarity that utilizes BERT's pretrained contextual embeddings. BERTScore is calculated by maximizing pairwise embedding similarities for the tokens of an original and its obfuscated text. All scores lie in $[0, 1]$, and higher scores denote greater semantic similarity.
    \item \textbf{Fluency.} Lastly, we measure the perplexity of obfuscated texts to ensure that the obfuscation process does not diminish the human readability of obfuscated texts. The perplexity is calculated as the negative log-likelihood of LLaMA2-7B~\cite{touvron2023llama} over obfuscated texts.
\end{itemize}

\section{Results}

\noindent \textbf{Obfuscation Success.} The experimental results on both datasets from our main obfuscation experiment are summarized by Table~\ref{table:attk_success_and_sim}. In the table, we denote the metric indicating the most favorable attack in bold (the metric with the lowest magnitude for obfuscation success metrics, and the metric with the highest magnitude for semantic preservation metrics) across each adversarial trial. Additionally, for the rows containing results for {\n}, we show the percentage change of each metric from the method that was the highest performing, excluding {\n}. Therefore a lower percentage (higher degradation of adversarial accuracy / F1-Score) is more desirable for obfuscation success metrics, while a higher percentage (less semantic degradation) is favorable for semantic preservation metrics.

On TuringBench, we see that {\n} is consistently the best performer in terms of attack success. {\n} consistently degrades adversarial accuracy more than other methods, demonstrating improvement as high as $21.90\%$. Additionally, F1-Score even more pronounced degradation, with improvement as high as $23.93\%$.

On the Blog Authorship Corpus, results shown in Table~\ref{table:attk_success_and_sim} indicate that {\n} is consistently the best performer in terms of F1-Score and accuracy. 

\noindent \textbf{Ablation of Interpretability-Based Replacement.} We observe that {\n} outperforms TextFooler-POS in all trials. This demonstrates the value of {\n}'s sequence replacement schema and interpretability-centric approach when compared to traditional token-by-token perturbation methods.

\noindent \textbf{Computational Complexity.} Running time results are summarized by Table~\ref{table:runningtime}. The One-Time Training stage encompasses all operations associated with data feature extraction and one-time training, while Inference corresponds to per-text running time.

\begin{table}[tb]
    \centering
    \footnotesize
    \begin{tabular}{lccc}
        \toprule
        \multicolumn{1}{l}{\textbf{Method}}   & \textbf{One-Time Training} & \textbf{Inference} \\ \hline
        \multicolumn{3}{c}{\textbf{TuringBench}} \\ \hline
        \multicolumn{1}{l}{Mutant-X} & 4 hrs & 3 min \\
        \multicolumn{1}{l}{Avengers} & 6 hrs & 5 min \\    
        \multicolumn{1}{l}{TextFooler-wordCNN\*} & 2 hrs & 8 sec \\
        \multicolumn{1}{l}{TextFooler-wordLSTM\*} & 2 hrs & 7 sec \\
        \multicolumn{1}{l}{BERT-Attack\*} & 6 hrs & 8 sec \\
        \multicolumn{1}{l}{{\n}} & \textbf{12 min}   & \textbf{0.8 sec}  \\ \hline
        \multicolumn{3}{c}{\textbf{Blog Authorship Corpus}} \\ \hline
        \multicolumn{1}{l}{Mutant-X} & 8 min & 10 min \\
        \multicolumn{1}{l}{Avengers} & 24 min & 14 min \\
        \multicolumn{1}{l}{TextFooler-wordCNN\*} & 2 hrs  & 11 sec \\
        \multicolumn{1}{l}{TextFooler-wordLSTM\*} & 2 hrs & 9 sec \\
        \multicolumn{1}{l}{BERT-Attack\*} & 6 hrs & 9 sec \\
        \multicolumn{1}{l}{{\n}} & {\bf 6 min} & \textbf{1.0 sec} \\ 
        \bottomrule
        \end{tabular}
    \caption{Statistics of the one-time training runtime and the average inference time per one sample for all methods.}
    \label{table:runningtime}
\end{table}

The results indicate that {\n} outperforms all baselines both in terms of one-time training and obfuscation runtime. {\n}'s total time for both one-time training and obfuscation of 100 samples indicates at least a 10x speed-up on TuringBench and at least an 18x speed-up on the Blog Authorship Corpus. {\n} is additionally at least 10x faster on TuringBench and 20x faster on the Blog Authorship Corpus with respect to one-time training and at least 10x faster during obfuscation on both datasets.


\vspace{3pt}
\noindent \textbf{Semantic Preservation.} Across both datasets, {\n} consistently outperforms in semantic preservation when evaluated with USE cosine similarity, the most robust measure of semantic preservation we measured, and BERTScore. However, we observe that {\n} consistently performs the worst in terms of METEOR score on both datasets; however, we believe that this result can largely be attributed to the inherent flaws of the METEOR score, as it is generally less correlated with human judgments when compared to USE cosine similarity, which is a stronger standard for semantic similarity analysis. We demonstrate these limitations in the Appendix.

\noindent \textbf{Fluency.}
Table ~\ref{tab:perplexity} demonstrates that {\n} demonstrates the best perplexity across both datasets, indicating the highest readability across all AO methods.

\begin{table}[h!]
\centering
\begin{tabular}{@{}lcc@{}}
\toprule
\textbf{Method}              & \textbf{TuringBench}    & \textbf{Blog}           \\ \midrule
Mutant-X            & 65.12          & 29.55          \\
Avengers            & 64.51          & 23.12          \\
TextFooler-wordCNN  & 57.69          & 17.96          \\
TextFooler-wordLSTM & 52.89          & 19.28          \\
TextFooler-POS      & 56.23          & 18.34          \\
{\n}                & \textbf{20.82} & \textbf{12.11} \\ \bottomrule
\end{tabular}
\caption{Perplexity of post-obfuscation texts measured using LLaMA2-7B (lower is better).}
\label{tab:perplexity}
\end{table}

\section{Discussion}
\noindent \textbf{Author Label Bias.} First, we analyze the distribution of author frequencies before and after obfuscation to identify potential obfuscation bias towards an author or set of authors on both datasets. To do this, we calculate the normalized entropy of author labels over obfuscated samples.



\begin{table*}[!ht]
\footnotesize
\centering
\begin{tabular}{lcccccc}
\toprule
\multirow{2}{*}{\textbf{Method}} &
  \multicolumn{2}{c}{\textbf{Obfuscation Success}} &
  \multicolumn{3}{c}{\textbf{Semantic Preservation}} \\ \cmidrule(lr){2-3} \cmidrule(lr){4-6}
&
  \textbf{Accuracy} $\downarrow$ &
  \textbf{F1-Score} $\downarrow$ &
  \textbf{METEOR} $\uparrow$ &
  \textbf{USE Cosine Similarity} $\uparrow$ &
  \textbf{BERTScore} $\uparrow$ \\ 
\midrule
GPT Output Detector - Base & 0.5000 & 0.3670 & 0.6966 & 0.8754 & 0.8941 \\
GPT Output Detector - Large & 0.5682 & 0.3623 & 0.6948 & 0.8734 & 0.9017 \\
GPTZero & 0.6170 & 0.5323 & 0.6897 & 0.8717 & 0.8936 \\
DetectGPT & 0.5729 & 0.4984 & 0.7478 & 0.9030 & 0.9134 \\
\bottomrule
\end{tabular}
\caption{{\n}'s attack success and semantic preservation against four machine text detection models.}
\label{table:chat_attk_and_sim}
\end{table*}

Because of the varied attack successes of different methods, we do not consider the raw entropy values but instead, consider the proportion of the total label entropy each author contributes. 
The distribution of these label entropy proportions should be as uniform as possible so that each author label transforms in an unpredictable way. A non-uniform entropy distribution across authors indicates that the obfuscation of a small pool of authors' texts contributes significantly to the overall attack success. This indicates a bias during obfuscation in regard to the transformation of author labels, a bias that can potentially be exploited by the attacked model. If the post-obfuscation prediction label were predictable based on the pre-obfuscation prediction label, an adversary would be able to gain significant information about the authorship of a text based on the predicted author post-obfuscation. This bias is further not desirable since the authorship pool may vary from various obfuscation settings.

\begin{figure}[!h]
    \centering
    \includegraphics[width=\columnwidth]{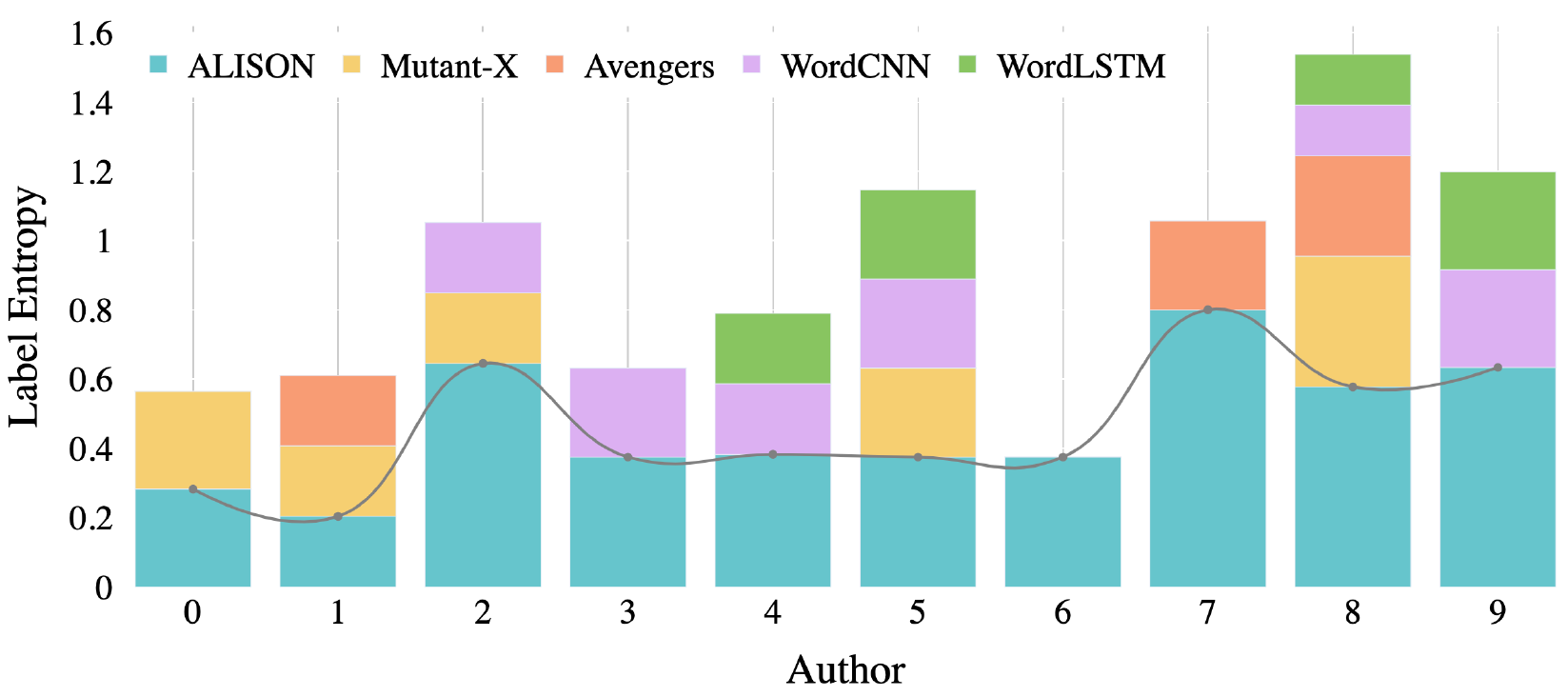}   \caption{Distribution of author-wise contributions to label entropy post-obfuscation.}
    \label{fig:EntropyALL}
\end{figure}

We present the individual author entropy contributions over all authors for all methods in Figure~\ref{fig:EntropyALL}. It is visually apparent that the distribution of author entropy contributions is significantly more uniform for {\n} when compared to other methods. This indicates significantly less predictability and label bias during obfuscation when compared to other methods. There are very few labels with a small or nonexistent contribution to overall entropy, which are labels that could be trivially reverse-engineered by the targeted model, unlike the entropy distributions of other methods.

\vspace{3pt}
\noindent \textbf{Interpretability.} 
Because {\n} relies on explicitly determined criteria for obfuscation, it can explain obfuscation decisions using quantified token importances. Interpretability is generated by extracting the POS n-grams in a text and using Integrated Gradients to generate the importance of each POS n-gram, which is scaled as described previously. Top POS n-gram features may then be mapped to specific token sequences in the original text.



\section{A Use Case: Obfuscating ChatGPT Texts}
The impressive performance of ChatGPT~\cite{gpt4}, a conversational language model, has led to its ubiquitous use in the workplace and classroom. Though ChatGPT can assist humans with everyday tasks, its potentially dishonest applications (e.g. construing ChatGPT's output as human-written text in academic settings) make the identification of ChatGPT-written texts an important problem with extensive commercial and academic study ~\cite{gptzero, mitchell2023detectgpt, gpt2release, upton}. The commercial value of ChatGPT detection further motivates an AO technique that is computationally efficient.

\vspace{3pt}
\noindent \textbf{Problem Formulation.} 
The real-world task of discriminating between ChatGPT and human-written texts is an increasingly relevant AA task that motivates the study of the corresponding AO task. We select four well-known machine-text generators, each demonstrating $>95\%$ discrimination accuracy, to study under adversarial perturbation: GPTZero ~\cite{gptzero}, DetectGPT ~\cite{mitchell2023detectgpt}, and both the Base and Large GPT Output detectors ~\cite{gpt2release} released by OpenAI.

\vspace{3pt}
\noindent \textbf{Methodology.} We used news article headlines from TuringBench to query the OpenAI Completions API. A single request was made for each unique headline, which consisted of a fixed generation prompt prepended to the headline. The corresponding human-written texts in the TuringBench corpus provided negative examples to introduce into the corpus, generating a set of evenly distributed negative and positive examples. The experimental setup described previously was then repeated.

\vspace{3pt}
\noindent 
\textbf{Main Obfuscation Trial Result.} Table~\ref{table:chat_attk_and_sim} shows metrics of Obfuscation Success and Semantic Preservation against adversarial classifiers. {\n} demonstrates degradation of adversarial accuracy to \emph{at most} $0.617$ and adversarial F1-Score to \emph{at most} $0.5323$. In addition, {\n} consistently maintains a high degree of semantic similarity between original and obfuscation texts, maintaining \emph{at least} $0.8717$ USE Cosine Similarity and $0.8936$ BERTScore. ChatGPT text detectors become negligibly useful at such adversarial performance, as the adversarial accuracy is close to the trivial accuracy of $0.50$ in the binary classification setting. 

\vspace{3pt}
\noindent
\textbf{Entropy Result.} We observe an entropy of 0.56 associated with the human class and an entropy of 0.44 associated with the ChatGPT class. Because the distribution of authorship label entropy is not significantly skewed toward any class, {\n} does not demonstrate a significant degree of bias during the obfuscation process in transferring attributions from any specific class.

\section{Conclusion}
We have presented a new authorship obfuscation technique, {\n}, based on the replacement of revealing stylistic sequences. {\n} greedily replaces text sequences matching POS n-grams identified to be important by interpreting a lightweight neural network trained to perform authorship attribution using mixed n-grams. We use {\n} to attack three SOTA transformer-based attribution classifiers and demonstrate an improvement in obfuscation success and semantic preservation when compared to \emph{seven} diverse baselines. We demonstrate that {\n}'s intuitive and simple but effective nature demonstrates a drastic improvement in computational complexity compared to baseline methods. Parameter analysis, qualitative analysis of {\n}'s obfuscated texts, limitations of METEOR score, etc. are presented in the Appendix.

\section*{Ethical Statement}

While authorship obfuscation enables freedom of speech for various previously described individuals including whistleblowers and journalists, it also potentially permits malicious groups to stay hidden. We acknowledge such ethical concerns but stress the need to study and design systems that can protect and enhance the freedom of speech of the public.

\section{Acknowledgements}
This work was supported in part by NSF awards \#1820609, \#1950491, and \#2131144.


\bibliography{aaai24}
\newpage

\appendix

\subsection{Hyperparameter Analysis} 

\begin{figure}[h!]
    \centering
    \includegraphics[width = 0.5\textwidth]{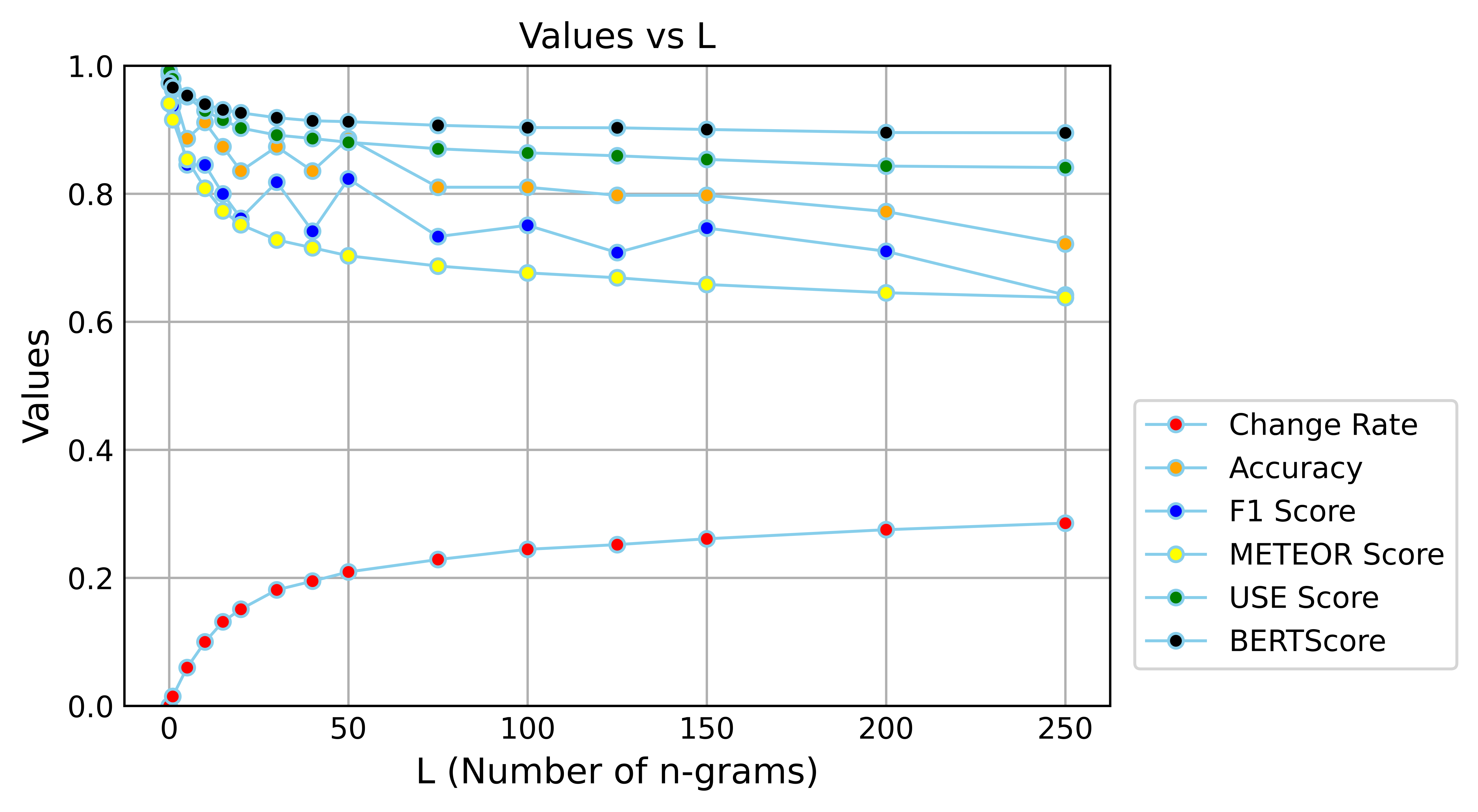}
    \caption{Effect of varying $L$ on obfuscation success and semantic preservation}
    \label{fig:valuesvl}
\end{figure}

In the following, we explore the effects of varying two key parameters on metrics of obfuscation on TuringBench.
We first explore the effects of varying the value of $L$. We record the previously described metrics for values of $L$ ranging from 1 to 250. The results of this experiment are summarized by Fig. ~\ref{fig:valuesvl}.

\noindent This figure indicates that increasing the value of $L$ increases the change rate and decreases all other metrics (both metrics of obfuscation success and semantic preservation). This result is intuitive, as increasing the number of sequences to change will, of course, increase the change rate and will therefore be more successful in fooling the adversarial classifier, and will also cause a greater difference between the original and obfuscated texts, decreasing semantic preservation. 

\medskip

\begin{figure}[!h]
    \centering
    \includegraphics[width = 0.5\textwidth]{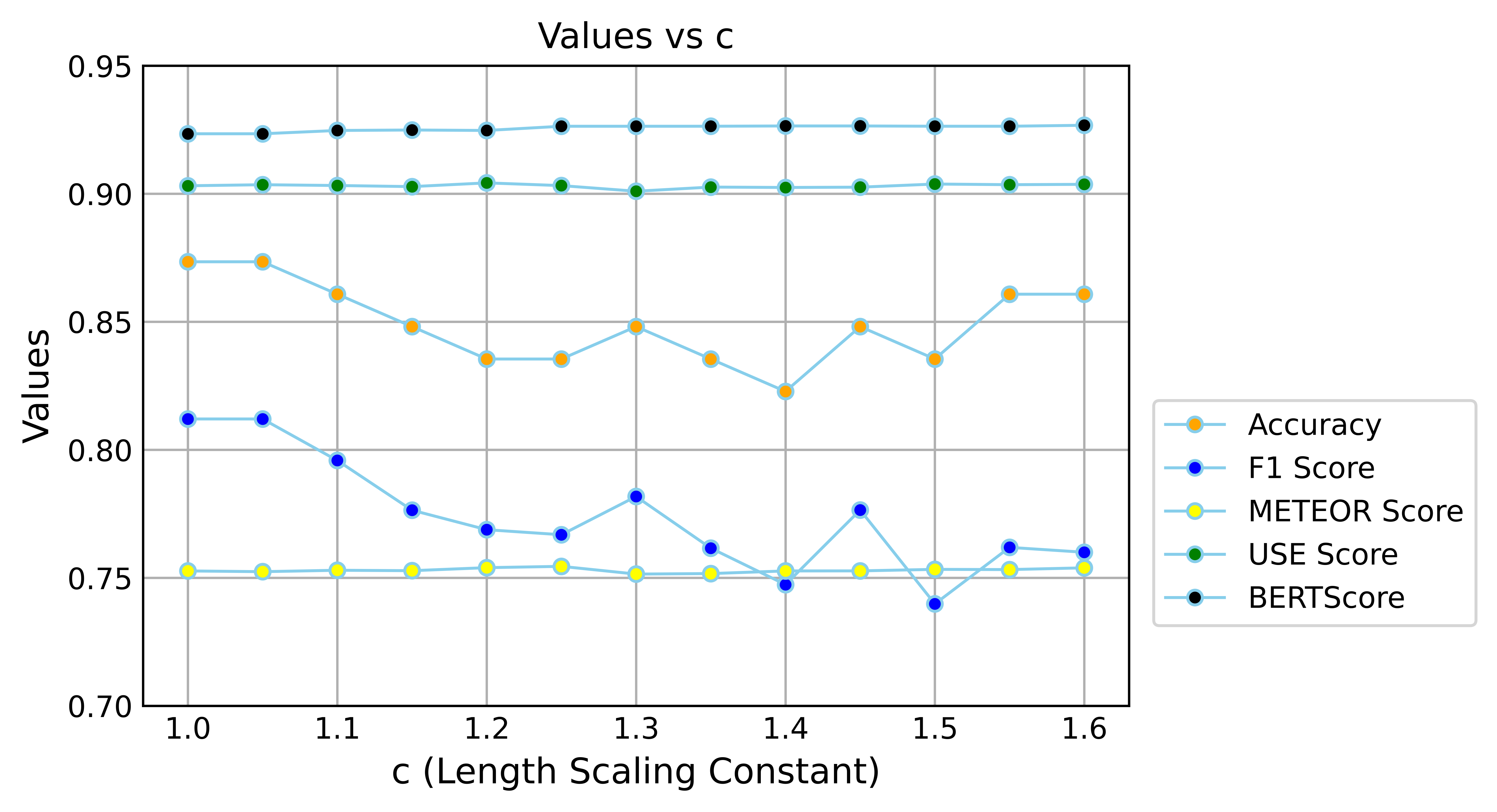}
    \caption{Effect of varying $c$ on obfuscation success and semantic preservation}
    \label{fig:valuesvc}
\end{figure}

We also explore the effect of varying the value of $c$ from 1.0 to 1.6. The results of this experiment are summarized in Fig. \ref{fig:valuesvc}.

\noindent From this, we observe that the value of $c$ does not significantly impact metrics of semantic preservation, as the adversarial METEOR, USE Cosine Similarity, and BERTScore do not significantly change across values of $c$. However, metrics of obfuscation success decrease, then increase across $c\in[1.0, 1.6]$, where the optimal value of $c$ is between 1.3 and 1.5. This supports the idea that artificially scaling importance based on length allows {\n} to choose sequences of optimal length to mask.

\subsection{Limitations of METEOR Score}
While {\n} generally outperforms other baselines, we demonstrate consistent underperformance when measuring semantic preservation in terms of METEOR score. We believe that this behavior is due to the inherent bias of the METEOR score against {\n}'s style of perturbation. First, we observe that the METEOR score highly depends on exact spacing. During the process of encoding and decoding associated with obtaining masked-phrase substitutions, erroneous spaces are introduced into the text. While these spaces do not impact information loss significantly, they do significantly impact the METEOR score. Without any phrase substitutions, the spacing changes made by PLM and the reconstruction process degrade the METEOR score to 0.93 on TuringBench and 0.90 on the Blog Authorship Corpus. However, the USE Cosine Similarity and BERTScore are minimally affected, maintaining values of around 0.99. An example of a text exhibiting near-perfect semantic preservation (as measured qualitatively, by USE Cosine Similarity, and by BERTScore) but with a low METEOR score is the following: 

\begin{itemize}
    \item \textit{Original}: That is what it is.   Cold and dreary.   A Soda without fizz. Boogers.
    \item \textit{Obfuscated}: i know what it is .Cold and dreary .A Soda without fizz .Boogers.
\end{itemize}

\noindent While the USE Cosine Similarity is 0.9409 and BERTScore is 0.9616, the METEOR score is 0.5152. The patterns exhibited in this example, leading to a low METEOR score, are repeated throughout many examples in our obfuscation method.

However, other baselines do not exhibit these patterns, with semantic preservation scores generally agreeing more. This is because single-word substitutions often leave spacing patterns and the relative ordering of words generally unchanged. The following, a sample taken from Avengers, illustrates this:

\begin{itemize}
    \item \textit{Original}: Went to an information session for people who might be interested in helping to teach the First Year Lawyering program next year... it's an organization called the Board of Student Advisors.  The meeting wasn't funny enough to give me anything funny to write.
    \item \textit{Obfuscated}: Went to an info meeting for children who might become interested in helping to teach the First Year Lawyering initiative next month... it's An watchdog called the Board of Student Advisors.  The press wasn't funny enough to get me nothing humorous to publish.   
\end{itemize}

\noindent This sample demonstrates a METEOR score of 0.7673, a USE Cosine Similarity of 0.8117, and a BERTScore of 0.9452.

\subsection{Qualitative Analysis}

We analyze the types of changes made by sampling some of the sentences pre- and post-obfuscation on TuringBench and the Blog Authorship Corpus. 
The changes generally follow a few distinct patterns:

\begin{itemize}
    \item Synonym Substitutions: Often, a word in the original text is replaced by a synonym in the obfuscated texts. For example, memo $\Longleftrightarrow$ novel, perfectly $\Longleftrightarrow$ well, smile $\Longleftrightarrow$ laugh, etc. These most likely result from a short POS-tag sequence being identified as important, leading to the PLM being able to relatively accurately determine the meaning of the masked words from context, leading to little information loss.
    \item Contextually Acceptable Substitutions: These substitutions involve the substitution of a word or phrase with a semantically unequivalent phrase that seems plausibly correct in context. For example, white $\Longleftrightarrow$ black, shirt $\Longleftrightarrow$ dress, shorted $\Longleftrightarrow$ given, etc. These most likely arise from a phrase being masked that cannot be reliably determined from the surrounding context. This leads to the PLM making an incorrect inference regarding the semantic meaning of the masked phrase. While this increases the degree of information loss between the original and obfuscated texts, it does not significantly affect the readability of the obfuscated text.
    \item Deletions: Sometimes, nonessential words and phrases are completely deleted after reconstruction with a PLM. 
    These deletions often occur in introductory phrases that contribute little semantic meaning. Although rarely happen, these deletions may slightly contribute to information loss or decrease readability.
    \item Equivalent Substitutions Resulting in Solecism: Sometimes, words or short phrases will be replaced by nearly identical substitutes semantically and functionally that result in a solecism. 
    This results in no information loss, however, the solecism may result in decreased readability. These substitutions may indicate inherent limitations of BERT for Masked Language Modeling, or inherent grammatical errors in the data BERT was trained on. Additionally, the uncapitalized "i" is also common.
\end{itemize}

\end{document}